\documentclass[conference]{IEEEtran}
\IEEEoverridecommandlockouts

\usepackage{cite}
\usepackage{amsmath,amssymb,amsfonts}
\usepackage{algorithmic}
\usepackage{graphicx}
\usepackage{textcomp}
\usepackage{xcolor}
\usepackage{booktabs}
\usepackage{url}
\def\BibTeX{{\rm B\kern-.05em{\sc i\kern-.025em b}\kern-.08em
    T\kern-.1667em\lower.7ex\hbox{E}\kern-.125emX}}
\begin{document}

\title{SNAP-FM: Sparse Nonlinear Accelerated Projection for Physics-Constrained
Generative Modeling\\

\thanks{This material is based upon work supported by the U.S. National Science Foundation under award Nos CNS-2346520,  RISE-2425761, and DMS-2325184, by the Defense Advanced Research Projects Agency (DARPA) under Agreement No. HR00112490488,  by the Department of Energy, National Nuclear Security Administration under Award Number DE-NA0004266
 and by the United States Air Force Research Laboratory under Cooperative Agreement Number FA8750-19-2-1000.  Neither the United States Government nor any agency thereof, nor any of their employees, makes any warranty, express or implied, or assumes any legal liability or responsibility for the accuracy, completeness, or usefulness of any information, apparatus, product, or process disclosed, or represents that its use would not infringe privately owned rights. Reference herein to any specific commercial product, process, or service by trade name, trademark, manufacturer, or otherwise does not necessarily constitute or imply its endorsement, recommendation, or favoring by the United States Government or any agency thereof.  The views and opinions of authors expressed herein do not necessarily state or reflect those of the United States Government or any agency thereof.}
}

\author{
\IEEEauthorblockN{
Alaina Kolli\textsuperscript{1,*},
Theodoros Xenakis\textsuperscript{1,2,*},
Utkarsh Utkarsh\textsuperscript{1},
Pengfei Cai\textsuperscript{1},\\
Rafael Gómez-Bombarelli\textsuperscript{1},
Alan Edelman\textsuperscript{1},
Christopher V. Rackauckas\textsuperscript{1,\textdagger}
}

\IEEEauthorblockA{
\textsuperscript{1}Massachusetts Institute of Technology,
Cambridge, MA, USA\\
\textsuperscript{2}Norwegian University of Science and Technology,
Trondheim, Norway\\
\textsuperscript{*}Equal contribution.\\
\textsuperscript{\textdagger}Corresponding author:
\texttt{crackauc@mit.edu}
}
}


\maketitle

\begin{abstract}
Generative models have emerged as scalable surrogates for physical simulation, yet they offer no guarantee that their outputs respect the conservation laws, boundary conditions, and nonlinear invariants that govern the underlying physics. Constrained sampling closes this gap, enforcing such constraints exactly at inference time without retraining, but at a computational cost: projection, correction and trajectory-optimization steps are repeated during sampling, with these steps becoming expensive for nonlinear constraints.
Standard ML frameworks exacerbate this: their dense tensor algebra and limited sparse solver composability obscure the structure that physical constraints naturally induce, making efficient batched nonlinear optimization difficult to realize in practice. We address this bottleneck by exploiting the structure that sample-wise batching and local PDE couplings induce in the projection subproblems -- namely, block-sparse Jacobian and KKT systems -- exposing this structure using \texttt{ExaModels.jl} and solving the resulting sparse nonlinear programs with \texttt{MadNLP.jl} and GPU sparse factorization. Applied to Physics-Constrained Flow Matching (PCFM), on PDE benchmarks with linear, nonlinear, one-dimensional, and two-dimensional constraints, this approach accelerates nonlinear constraint projection while maintaining constraint satisfaction. These results show that sparse GPU nonlinear optimization is a practical foundation for constrained generative sampling in scientific machine learning. 
\end{abstract}


\section{Introduction}

Generative models have emerged as flexible surrogates for physical
simulation, learning solution distributions for partial differential equations
(PDEs) and amortizing inference across varying physical conditions~\cite{price2023gencast,yuan2023physdiff,huang2024diffusionpde,
utkarsh2025pcfm}. Yet their deployment in scientific settings is limited by a
fundamental gap: unconstrained generative models do not, by themselves,
guarantee physical fidelity. Conservation of mass, momentum, and energy,
nonlinear boundary conditions, and invariants tied to the governing equations
are central to classical numerical simulation, but are routinely violated by
learned surrogates unless explicitly enforced~\cite{RAISSI2019686,li2021fourierneuraloperatorparametric}. Closing this gap
without sacrificing the amortized inference advantage of generative models is
the central challenge of physics-constrained generative modeling.

Constraint enforcement in generative sampling can be broadly divided into soft
and hard approaches. Soft methods, including training-time penalty losses~\cite{baldan2025flow,huang2024diffusionpde}, PINN-style residual regularization,
physics-informed neural operators, and architecture-level inductive biases~\cite{greydanus2019hamiltonian,richter2022neural},
encourage constraint satisfaction approximately. These approaches are
computationally attractive and scale naturally within modern deep learning
pipelines, but they generally provide no exact feasibility guarantees and can
exhibit increased constraint violation under distribution shift.

Hard-constrained methods instead aim to enforce feasibility exactly, either
through inference-time correction and optimization~\cite{utkarsh2025pcfm,christopher2024constrained,
cheng2025gradientfreegenerationhardconstrainedsystems,romer2024diffusion,
yuan2023physdiff} or through end-to-end constrained formulations~\cite{utkarsh2025end}. In the simplest case, one can apply a final post-hoc
projection that maps a generated sample onto the feasible manifold, but such a
late correction can introduce substantial distributional distortion. To reduce
this mismatch, many methods interleave corrections with the sampling dynamics,
repeatedly guiding or projecting intermediate states toward feasibility~\cite{utkarsh2025pcfm,
cheng2025gradientfreegenerationhardconstrainedsystems,
christopher2024constrained,benhamu2024dflowdifferentiatingflowscontrolled}.
Relaxed variants further delay or soften early corrections, reflecting the fact
that early iterates often remain noise-like and may be harmed by overly strict
constraint enforcement.

Despite these algorithmic differences, hard-constrained test-time methods share
a common computational burden: feasibility is enforced through repeated
optimization during sampling~\cite{utkarsh2025pcfm,
cheng2025gradientfreegenerationhardconstrainedsystems,
christopher2024constrained}. In projection-based methods, this burden is
especially explicit: each correction maps an intermediate state onto, or closer
to, the constraint manifold by solving a constrained optimization problem. For
linear constraints, this projection often reduces to a single matrix solve. For
the nonlinear conservation laws governing many physical systems, including flux
constraints, integral invariants, and nonlinear boundary conditions, the same
step requires iterative nonlinear optimization. Since this optimization must be
performed repeatedly across sampling steps and over batches of generated
samples, it can dominate the total cost of hard-constrained generation. The
central question is therefore not merely whether to enforce hard constraints,
but how to make the resulting projection and correction steps fast enough to be
practical at scale.

The algorithmic response is well known from large-scale optimization: exploit
the sparsity induced by the constraints, and solve the resulting KKT systems (see for example section~\ref{subsec:jacobian-structure})
using sparse linear algebra~\cite{numericalOptimization,pacaud2024gpu,rennich2014accelerating,
lu2025cupdlp}. However, this strategy is difficult to deploy inside modern
generative sampling pipelines. Standard ML frameworks are optimized primarily
for dense batched tensor algebra, and although sparse tensor support exists,
it does not yet provide a mature end-to-end stack for batched nonlinear
programming, interior-point globalization, GPU-resident sparse KKT
factorization, and integration with the sampling loop~\cite{paszke2019pytorch,bradbury2018jax}. This gap between constrained
optimization practice and deep generative modeling infrastructure motivates our
approach.

The projection NLPs arising in batched hard-constrained sampling are highly
structured. Across the batch, the constraint Jacobian is block diagonal because
each sample's constraints are independent of the others. Within each block, the
Jacobian is sparse because physical conservation laws and PDE discretizations
couple only local spatial or temporal degrees of freedom. This two-level
structure yields sparse KKT systems that are amenable to GPU-resident sparse
factorization, avoiding the dense tensorization that would otherwise make
nonlinear projection prohibitively expensive.

We realize this idea through a Sparse Nonlinear Accelerated Projection Flow-Matching (SNAP-FM) framework, based on Physics-Constrained Flow Matching (PCFM)~\cite{utkarsh2025pcfm}. SNAP-FM uses \texttt{ExaModels.jl} to symbolically
compile the structured projection NLP, \texttt{MadNLP.jl} as the interior-point
solver, and GPU sparse factorization for the resulting KKT systems. Built on
PCFM, which enforces arbitrary nonlinear constraints zero-shot in pretrained
flow models, SNAP-FM targets the main computational bottleneck of
constraint enforcement: repeated nonlinear projection during sampling. Across
heat, reaction-diffusion, Burgers, and two-dimensional Navier-Stokes
benchmarks, SNAP-FM accelerates the projection step relative to generic optimization baselines while maintaining constraint
satisfaction.

\section{Physics Constrained Flow Matching}
This work builds directly on the Physics-Constrained Flow Matching framework introduced by Utkarsh et al.~\cite{utkarsh2025pcfm}, with the goal of extending it to a domain the current implementation doesn't support: the efficient enforcement of nonlinear constraints at scale. Among recent constraint-aware generative models, PCFM is the natural foundation for our work, also showing promising extension to different scientific domains such as atomistic generative models~\cite{cai2026enforcing}. It ensures constraint satisfaction exactly at inference time, requiring no retraining or architectural modifications of the underlying flow model, operating entirely post-hoc. It projects intermediate flow states onto constraint manifolds at inference time, without requiring gradient information during training, enabling zero-shot constraint enforcement consistent with prescribed solver tolerance for pretrained flow matching models. Unlike prior zero-shot approaches restricted to linear or non-overlapping constraints, it admits arbitrary nonlinear and coupled constraints within a single framework. These characteristics make PCFM a solid foundation on which to build a more general-purpose constrained sampling mechanism. 

\subsection{Sampling}\label{sec:Sampling}
The sampling algorithm utilized in PCFM is a constraint-guided algorithm that interleaves lightweight constraint corrections with marginally consistent flow updates. The procedure consists of four main steps: forward shooting, Gauss-Newton projection, reverse updating, and relaxed constraint corrections. 

Let $v_\theta(u, \tau)$ denote the pretrained flow model, which defines an ODE transporting samples $u_0 \sim \pi_0$ from a tractable prior to solution-like outputs $u_1 \sim \pi_1$ over flow time $\tau \in [0, 1]$. Given a constraint function $h(u)$, PCFM enforces $h(u_1) = 0$ by discretizing $[0, 1]$ into $N$ uniform substeps. At each timestep, $\tau \rightarrow \tau' = \tau + \Delta \tau$, three operations are performed: a one-step extrapolation to the terminal time using the learned velocity, a projection of the resulting candidate onto the constraint manifold $\mathcal{M} = \{u : h(u) = 0\}$, and a linear interpolation back to $\tau'$ along the optimal-transport (OT) interpolant~\cite{utkarsh2025pcfm}:
\begin{align}
  \hat{u}_1     &= u_\tau + (1 - \tau)\, v_\theta(u_\tau, \tau), \\
  u_1           &= \arg\min_u\; \tfrac{1}{2}\|u - \hat{u}_1\|^2
                   \quad \text{s.t.} \quad h(u) = 0,
                   \label{eq:projection} \\
  u_{\tau'}     &= u_0 + \tau'(u_1 - u_0).
                   \label{eq:ot-pullback}
\end{align}
The full PCFM algorithm of~\cite{utkarsh2025pcfm} additionally permits a penalized correction at each step, $\arg\min_u \|u - \hat{u}_{\tau'}\|^2 + \lambda \|h(u + (1-\tau')v_\theta(u,
\tau'))\|^2$, to compensate for nonlinearity under coarse discretization. However, we omit it, as in our experiments, the streamlined three-step variant is sufficient.

The key structural feature exploited in this work is that the loop above is constraint-agnostic. Every operation except the projection in step 2 is a fixed arithmetic kernel, learning the entire specifications of the problem to be funneled into the projection subproblem. The cost of a single sampling pass, therefore, can be divided into two main terms: the constraint projections and the velocity evaluations. The forward shoot and OT pullback in eq.~\eqref{eq:ot-pullback} are independent of the constraints being enforced, whereas the projection step scales with the number of imposed constraints and their complexities, causing the projection step to be the bottleneck of the sampling, especially in these more complex cases. The remainder of this work focuses on optimizing the projection step by exploiting the sparsity in the Jacobians, since constraints are independent across samples, and temporal couplings are typically local.  

\section{Theory}
In this section, we discuss common constraints imposed on PDE systems and derive structural properties of the associated constraint Jacobians. 

We adopt notation similar to that of Utkarsh et al.~\cite{utkarsh2025pcfm}. Specifically, we consider partial differential equations on a bounded spatiotemporal domain $\Omega \times [0,T]$, where $ \Omega \subset \mathbb{R}^d$.
The true PDE solution is a field $u: \Omega \times [0,T] \to \mathbb{R}$, which we approximate using a discretized tensor $\mathbf{u}\in \mathbb{R}^{N_x\times N_t}$, where $N_x$ and $N_t$ are the number of spatial and temporal grid points, respectively.

In addition to satisfying the governing PDE, the solution must also satisfy a collection of physical constraints. These constraints are represented as \begin{align*}
    (\mathcal{H}_k(u))_k = \mathbf{0}
\end{align*}
where $\mathcal{H}_k$ is the $k$-th constraint acting on the solution $u$.
For convenience, we let $\mathcal{H}(u)$ denote the concatenation of all imposed constraints. Further, it is implied 
that the constraint operator can also work on the discretized solution $\mathbf{u}$. 

Although the discussion in this section focuses on one-dimensional PDEs, the main ideas extend naturally to higher-dimensional settings.

\subsection{Constraints}\label{subsec:constraints}
Constraints imposed for PDEs typically arise from physical principles such as conservation laws, initial conditions, and boundary conditions~\cite{leveque1992FVM, leveque1992numerical}.
These constraints may be linear or nonlinear in $u$, and may act locally or globally in space and time. 
A selection of commonly encountered constraints is summarized in table \ref{tab:constraintTypes}.

\begin{table}[t]
  \caption{Overview of common constraint types for PDEs.}
  \label{tab:constraintTypes}
  \begin{center}
    \begin{small}
      \begin{sc}
        \begin{tabular}{lcccr}
          \toprule
          Constraint Type  & Form    \\
          \midrule
            Dirichlet IC/BC & $Au - b$  \\
            Global mass conservation& $\int_\Omega u \, dx - C$  \\
            Nonlinear conservation law& $\frac{d}{dt}\int_\Omega \rho(u) \, dx - C$\\
            Neumann or Flux BC& $\partial_n u(x,t) - g(x,t)$ \\
          \bottomrule
        \end{tabular}
      \end{sc}
    \end{small}
  \end{center}
  \vskip -0.1in
\end{table}

Most constraints encountered in practice are local in time. Meaning, they depend only on a single time step or a small number of neighboring time steps. For instance, an initial value constraint applies only at $t=0$, while a mass conservation constraint typically applies to each time step independently.
Even nonlinear conservation laws involving temporal derivatives can often be discretized using finite-difference schemes that couple only a small number of neighboring time steps, thereby preserving temporal locality~\cite{numericalMathematics}. 
The general Dirichlet IC/BC constraint formulated as $Au - b$ in table~\ref{tab:constraintTypes} could, in 
principle, define a global coupling, depending on the structure of $A$. In practice, however, $A$ is usually sparse and only couples nearby spatial or temporal nodes.

\subsection{Jacobian Structure}\label{subsec:jacobian-structure}
The Gauss-Newton projection step presented in section~\ref{sec:Sampling} can be reformulated as solving the constrained optimization problem 
\begin{align}
    u_{\text{proj}} = \arg\min_{u}\frac12||u-\hat{u}_1||_2^2\quad\text{subject to } \mathcal{H}(u) =0
\end{align}
where the variables $u$ can be interpreted as flattened vectors. Solving such a constrained optimization problem generally involves repeatedly solving large linear systems derived from the Karush-Kuhn-Tucker (KKT) conditions~\cite{numericalOptimization}. These systems typically take the form

\begin{align}
        \begin{bmatrix}
        W & J_{\mathcal{H}}^T\\
        J_{\mathcal{H}} & 0 
    \end{bmatrix}
    \label{eq:KKTsystem}
\end{align}
where $J_{\mathcal{H}}=\frac{\partial \mathcal{H}}{\partial u}$ is the Jacobian of the constraints $\mathcal{H}$, and  $W$ is the Hessian of the Lagrangian. We defer a detailed discussion of the KKT derivation to Appendix \ref{appdx:newtonKKT}. 
Regardless of the specific optimization method employed, computing the Jacobian - or an approximation to it - is generally a necessary part of the solution procedure~\cite{numericalOptimization}. 

A key observation is that physically motivated PDE constraints typically induce highly sparse Jacobians after discretization. Exploiting this sparsity is crucial for efficiently solving the associated linear systems. Moreover, for a fixed set of constraints, the sparsity pattern remains constant throughout the optimization process, allowing repeated linear solves to reuse structural information. 

To understand why $J_{\mathcal{H}}$ is sparse, consider the discretized solution as a 
flattened vector $\mathbf{u}\in \mathbb{R}^{N_x N_t}$. The Jacobian then has dimensions $J_{\mathcal{H}}\in \mathbb{R}^{N_c \times N_x N_t}$, where $N_c$ denotes the number of constraints.

If a constraint depends only on a single time step, at most $N_x$ entries of the corresponding Jacobian row will be nonzero. Similarly, if a constraint involves only a few spatial steps, but is global in time, then each row contains only a small number of nonzero entries per time step, yielding at most $\mathcal{O}(N_t)$ non-zero entries overall. 
Furthermore, constraints that are global in one domain but local in the other often induce repeated block structures in the Jacobian, since the same constraint operator is applied repeatedly across spatial or temporal points. 
For an involved example involving both mass and an initial value constraint, we refer to Appendix \ref{appdx:toyConstraints}.


It is also useful to consider the case where a batch of $N_s$ samples is generated simultaneously. In that setting, the Jacobian has dimension $J_{\mathcal{H}}\in \mathbb{R}^{N_c N_s \times N_x N_t N_s}$. Since each sample is independent, the resulting Jacobian exhibits a block-diagonal structure, with each block corresponding to the constraints of an individual sample.

\subsection{Sparse Solvers and GPU acceleration}
Several optimization frameworks have recently been developed to exploit the structure and sparsity of the objective and constraint functions in large-scale nonlinear programs (NLP). Two notable libraries are \texttt{ExaModels.jl}~\cite{shin2024examodels} and \texttt{MadNLP.jl}~\cite{shin2024madnlp}. 

ExaModels is an algebraic modeling and automatic differentiation framework designed for high-performance nonlinear optimization~\cite{shin2024examodels}. When using a general-purpose modeling framework such as \texttt{JuMP.jl}~\cite{Lubin2023}, it 
has been found that automatic differentiation may constitute a significant fraction of the total solver time~\cite{shin2024madnlp}. 
ExaModels is designed to mitigate this issue by generating sparsity-aware computational graphs and compiling them into SIMD-parallel kernels through \texttt{ExaCore}. This allows objectives, constraints, Jacobians and Hessians to be evaluated efficiently in a single sparsity-preserving pass, while exploiting the parallelism available on modern GPU architectures~\cite{shin2024madnlp}.

MadNLP implements a primal-dual interior point solver, working together with ExaModels to form and solve sparse KKT systems efficiently. Since the sparsity pattern of the Jacobian typically remains fixed throughout the optimization process, MadNLP can leverage this to optimize the linear system solves, significantly reducing 
overhead associated with repeated factorizations of the Jacobian~\cite{shin2024madnlp}.

Furthermore, MadNLP works across both CPU and GPU. On GPUs, MadNLP interfaces with 
NVIDIA's GPU sparse linear solver, cuDSS \cite{cudss}. cuDSS solves sparse linear systems using variants
of $LU$, $LDL^T$ or Cholesky factorization, and is specifically designed for exploiting 
the parallelism of NVIDIA GPU architectures. Large, sparse KKT systems can thus be solved
substantially faster than with traditional CPU-based solvers~\cite{pacaud2024gpu}.

\section{Methodology and Setup}

We evaluate SNAP-FM via benchmarking of the PCFM sampling loop across a set of PDE problems chosen to span the constraint regimes of practical interest, and for each problem, compare the runtime and constraint-satisfaction behavior of numerous optimization backends inside the projection step. For implementation details and scripts used to reproduce the experiments, we provide the code at \url{https://github.com/xenakistheo/PCFM.jl}. Additional details on the experimental setup, hardware, and benchmarking procedure are given in Appendix \ref{appdx:implementationDetails}.

\subsection{Optimization Methods}\label{subsec:methods}
Our main proposed solver, \textbf{ExaModels + MadNLP on GPU} is benchmarked against five baselines that utilize different algebraic modeling layers, NLP solvers, and execution hardware. We chose this set of methods such that comparisons against the adjacent baselines allow for isolation of contributions of each of the three axes.


\paragraph{ExaModels + MadNLP (GPU)}
As mentioned earlier, \texttt{ExaModels.jl}~\cite{shin2024examodels} compiles the projection NLP into SIMD-parallel GPU kernels via \texttt{ExaCore}, allowing for evaluation of objectives, constraints and AD in a single sparsity-preserving pass. The resulting structured KKT system is then solved entirely on the GPU by \texttt{MadNLP.jl}~\cite{shin2024madnlp}, an interior-point (IP) solver, with the sparse linear systems broken down via cuDSS. This combination is a promising direction for large-scale batches where the full Jacobian structure can be exploited for GPU-accelerated NLP. 

\paragraph{ExaModels + MadNLP (CPU)}
This optimization method is of identical modeling to our main method, but instead executed on the CPU. This baseline isolates the contribution of GPU acceleration. Namely, any performance gap that is found between this method and our above approach is purely attributed to hardware utilization.

\paragraph{JuMP + MadNLP (CPU)}
We utilize the same NLP solver, but with the model instead expressed in $\texttt{JuMP.jl}$~\cite{Lubin2023}, a general-purpose algebraic modeling package. JuMP differs from ExaModels in that it does not preserve SIMD structure or compile to GPU kernels. Therefore, this allows for isolation of the contribution of the modeling layer, specifically the value of structure-preserving compilation. 

\paragraph{JuMP + Ipopt}
Ipopt~\cite{Wachter2006Ipopt} is one of the most widely deployed NLP solvers in scientific computing. We include it as a standard benchmark any practitioner would reach for when including nonlinear constraints in \texttt{PCFM.jl}, and serves as empirical evidence that projection remains the sampling bottleneck in the nonlinear regime.  

\paragraph{Optimization.jl + IPNewton}
We implemented Optimization.jl with IPNewton to benchmark with a second interior-point baseline through the \texttt{Optimization.jl}~\cite{dixit2023optimization} unified interface. This ensures that any observed performance improvements are attributed correctly to MadNLP implementations of interior-point methods, not just the IP method itself.

\paragraph{Optimization.jl + L-BFGS}
We include L-BFGS as an unconstrained optimization baseline, ensuring the structured-NLP framing is truly advantageous and necessary.  Unlike the other methods, L-BFGS does not impose hard equality constraints. Instead, the constrained projection problem is converted into a different mathematical problem in which the constraint residuals enter through a penalty term. This distinction is important when interpreting runtime and feasibility results, since a faster L-BFGS should not be read as a faster solution of the hard-constrained projection problem. 

\subsection{Test Problems}\label{subsec:modelProbs}
We consider generative modeling for a collection of six different PDE benchmark problems with varying physical constraints. These problems are chosen to evaluate how different constraint sets influence both inference quality and computational cost. More importantly, we are also interested in how these results vary across the different solvers. The benchmark problems are as follows 

\begin{itemize}
    \item One dimensional heat equation with initial value and mass-conservation constraints.
    \item One dimensional heat equation with initial value, mass-conservation, and energy evolution constraints.
    \item One dimensional reaction diffusion equation with Neumann boundary conditions and a nonlinear conservation constraint. 
    \item One dimensional Burgers' equation with boundary and mass-conservation constraints. 
    \item One dimensional Burgers' equation with initial value, mass-conservation, and local flux constraints. 
    \item Two-dimensional Navier-Stokes equation in vorticity form with initial-value, total-vorticity conservation, and enstrophy constraints.
\end{itemize}

The test problems with their constraints are described in detail in Appendix \ref{appdx:ModelProblems}.

For both the heat equation and Burgers' equation, the paired test problems use identical pretrained functional flow matching (FFM) models~\cite{kerrigan2023functionalflowmatching}. In each case, the training dataset and network architecture are unchanged, and only the inference-time constraints differ. 

For the Burgers' with initial value constraint, we also experiment with removing the mass-conservation, and the flux constraints in a scaling study.


The discretization procedures used to enforce the constraints are described in Appendix \ref{appdx:constraintMotivation}.

\section{Experiments and Results}

We evaluate the proposed methods on the benchmark problems outlined in Section~\ref{subsec:modelProbs}.  These problems span constraint regimes of increasing difficulty, including linear and nonlinear one-dimensional constraints, as well as a nonlinear two-dimensional problem. For each benchmark, we compare all optimization backends from Section~\ref{subsec:methods} in terms of runtime, and constraint violation. 
Beyond the main benchmark results, we include two scaling studies to better characterize the computational behavior of the projection step. First, we consider the heat equation with fixed initial condition and mass-conservation constraints, and vary the number of generated samples to study scaling with batch size. Second, we consider the Burgers equation under increasingly rich constraint sets, allowing us to assess how solver performance changes as additional nonlinear constraints are imposed. The scaling studies are described in more detail in Appendix \ref{appdx:additionalResults}.

Unless otherwise stated, all reported runtimes are averaged over four repetitions with $32$ generated samples — an exception being the two-dimensional Navier-Stokes problem for which we generated $2$ samples due to the substantially larger state dimension. Wall-clock timings are reported using \texttt{BenchmarkTools.jl}'s \texttt{@btime} $\pm$ a standard deviation. \textit{DNF} denotes runs that did not converge within the allotted runtime budget of six hours.

\subsection{Runtime}
Tables \ref{tab:heatEquation1}, \ref{tab:heatEquation2}, \ref{tab:burgersBC}, \ref{tab:burgersIC}, \ref{tab:reactionDiffusion} \&  \ref{tab:NavierStokes} present the central results of this paper: end-to-end sampling time for a full pass of PCFM under each method and constraint regime. The fastest solver in each constraint regime is highlighted in bold.

\begin{table}[t]
  \caption{Runtime: Heat Equation - 1}
  \label{tab:heatEquation1}
  \begin{center}
    \begin{small}
      \begin{sc}
        \begin{tabular}{lccr}
          \toprule
          Optimization Setup  &  Time (s)  \\
          \midrule
          \textbf{Exa, MadNLP, GPU}     &   $\mathbf{11.73\pm 0.34}$ \\
          Exa, MadNLP, CPU     &   $14.29\pm0.15$   \\
          JuMP, MadNLP, CPU         &   $137.94\pm4.42$  \\
          JuMP, Ipopt, CPU          &   $182.72\pm 4.73$ \\
          IPNewton &   $6225.35\pm38.60$ \\
          L-BFGS   &   $1623.98\pm107.94$  \\
          \bottomrule
        \end{tabular}
      \end{sc}
    \end{small}
  \end{center}
  \vskip -0.1in
\end{table}

\begin{table}[t]
  \caption{Runtime: Heat Equation - 2}
  \label{tab:heatEquation2}
  \begin{center}
    \begin{small}
      \begin{sc}
        \begin{tabular}{lccr}
          \toprule
          Optimization Setup  &  Time (s) \\
          \midrule
          \textbf{Exa, MadNLP, GPU     }&    $\mathbf{110.26 \pm 1.92}$  \\
          Exa, MadNLP, CPU     &    $3501.11\pm 89.36$   \\
          JuMP, MadNLP, CPU         &    DNF  \\
          JuMP, Ipopt, CPU          &    DNF  \\
          IPNewton &    DNF  \\
          L-BFGS   &    $977.87\pm39.22$ \\
          \bottomrule
        \end{tabular}
      \end{sc}
    \end{small}
  \end{center}
  \vskip -0.1in
\end{table}

\begin{table}[t]
  \caption{Runtime: Burgers' (BC)}
  \label{tab:burgersBC}
  \begin{center}
    \begin{small}
      \begin{sc}
        \begin{tabular}{lccr}
          \toprule
          Optimization Setup  &  Time (s)  \\
          \midrule
          Exa, MadNLP, GPU     &   $153.53\pm 1.50$    \\
          \textbf{Exa, MadNLP, CPU}     &   $\mathbf{109.74\pm7.67}$   \\
          JuMP, MadNLP, CPU         &   $438.26\pm38.94$   \\
          JuMP, Ipopt, CPU          &   $1177.00\pm75.58$   \\
          IPNewton &   DNF   \\
          L-BFGS   &  $170.04\pm 21.89$   \\
          \bottomrule
        \end{tabular}
      \end{sc}
    \end{small}
  \end{center}
  \vskip -0.1in
\end{table}

\begin{table}[t]
  \caption{Runtime: Burgers' (IC, Mass, Flux)}
  \label{tab:burgersIC}
  \begin{center}
    \begin{small}
      \begin{sc}
        \begin{tabular}{lccr}
          \toprule
          Optimization Setup  &  Time (s) \\
          \midrule
          \textbf{Exa, MadNLP, GPU }    &   $\mathbf{38.83\pm0.55}$     \\
          Exa, MadNLP, CPU     &   $2048.83\pm187.58$  \\
          JuMP, MadNLP, CPU         &   DNF   \\
          JuMP, Ipopt, CPU          &   DNF    \\
          IPNewton &   DNF   \\
          L-BFGS   &   DNF    \\ 
          \bottomrule
        \end{tabular}
      \end{sc}
    \end{small}
  \end{center}
  \vskip -0.1in
\end{table}

\begin{table}[t]
  \caption{Runtime: Reaction-Diffusion Equation}
  \label{tab:reactionDiffusion}
  \begin{center}
    \begin{small}
      \begin{sc}
        \begin{tabular}{lccr}
          \toprule
          Optimization Setup  &  Time (s)  \\
          \midrule
          Exa, MadNLP, GPU     &   $35.68\pm 0.74$   \\
          \textbf{Exa, MadNLP, CPU}     &   $\mathbf{24.58\pm 1.69}$   \\
          JuMP, MadNLP, CPU         &   $173.15\pm 3.56$  \ \\
          JuMP, Ipopt, CPU          &   $228.80\pm4.35$  \\
          IPNewton &   $2314.20\pm62.47$   \\
          L-BFGS   &   $484.19\pm45.62$   \\
          \bottomrule
        \end{tabular}
      \end{sc}
    \end{small}
  \end{center}
  \vskip -0.1in
\end{table}

\begin{table}[t]
  \caption{Runtime: Navier-Stokes' Equation}
  \label{tab:NavierStokes}
  \begin{center}
    \begin{small}
      \begin{sc}
        \begin{tabular}{lccr}
          \toprule
          Optimization Setup &  Time (s)  \\
          \midrule
          \textbf{Exa, MadNLP, GPU}     &   $\mathbf{45.69\pm0.96}$   \\
          Exa, MadNLP, CPU     &   $118.49\pm0.84$   \\
          JuMP, MadNLP, CPU         &   $79.20\pm2.02 $   \\
          JuMP, Ipopt, CPU          &   $59.38\pm0.74$  \\
          IPNewton &   DNF   \\
          L-BFGS   &   $298.30\pm 14.26$    \\
          \bottomrule
        \end{tabular}
      \end{sc}
    \end{small}
  \end{center}
  \vskip -0.1in
\end{table}

A few observations worth highlighting.
Notably, the ExaModels + MadNLP combination beat the other methods in all six test problems. Moreover, the gap between the GPU and CPU implementation isolates the contribution of the hardware. This difference between the two was especially pronounced in the case of the heat equation with an energy-evolution constraint, as well as the Burgers' PDE with Godunov's flux constraints. We hypothesize that the nonlinearity of these two sets of constraints made it significantly harder to solve as an optimization problem. Hence, the difference in performance became more pronounced. 
For the smaller or less nonlinear constraint sets (e.g., \ref{tab:burgersBC},~\ref{tab:reactionDiffusion}), the CPU variant occasionally edges out the GPU implementation, as fixed kernel-launch and host-device transfer overheads are not yet amortized by problem size, becoming favorable only once the projection subproblem is large or nonlinear enough.

Secondly, the difference in the results of JuMP + MadNLP and ExaModels + MadNLP both run on CPU isolates the modeling layer, which is our main focus here. What we observe is that there is a consistent speedup from the exploitation of the sparsity structure of the Jacobian matrices at compile time. Even for the simple constraints in the heat equation, \ref{tab:heatEquation1}, the difference is significant. 

We also note that missing entries correspond to runs that did not complete within the runtime budget. These failures are themselves informative: all methods except for the ExaModels + MadNLP failed to converge on the Burgers' Equation constraint that included IC and flux constraints within a reasonable amount of time. IPNewton also failed on Burgers' with boundary constraints, as well as on the Navier-Stokes' equation.

\subsection{Constraint Satisfaction}
Beyond analyzing the runtimes, we want to verify that the optimization methods formulated as constrained nonlinear programs actually solve the stated projection problem. 
Since our primary objective is to compare
computational performance, we do not use small differences in constraint
residuals as an additional performance metric once feasibility has been
achieved to the prescribed numerical tolerance.

Across all benchmark problems, the constrained formulations based on
\texttt{ExaModels + MadNLP} and \texttt{JuMP} with either
\texttt{MadNLP} or \texttt{Ipopt} attain small constraint residuals
consistent with the prescribed solver stopping criteria. This confirms
that the reported runtime differences are not obtained by relaxing or
neglecting the imposed constraints.
As expected, \texttt{IPNewton} and \texttt{L-BFGS} do not attain the same
level of constraint satisfaction in our implementations.

A definition of the constraint-violation metric, together with the
complete results for every benchmark, is provided in
Appendix~\ref{appdx:additionalResults}.

\section{Conclusion}
We present an acceleration of the projection subproblem in Physics-Constrained Flow Matching by exploiting the sparsity structure inherent in the constraints. This work is motivated by the observation that the computational complexity of the projection step scales poorly during the PCFM sampling process as the number and complexity of constraints increase. By analyzing the Jacobian sparsity pattern, we reformulate the problem as a collection of independent per-sample subproblems while additionally leveraging the fixed sparsity structure arising from the discretized conservation constraints.
To exploit this structure efficiently on the GPU, we employed \texttt{ExaModels.jl} as the modeling layer together with \texttt{MadNLP.jl} as the nonlinear programming solver. Across our benchmarks, this approach yields a significant speedup for computationally demanding nonlinear constraints, while not compromising on correctness.

\section*{Acknowledgments}
The authors acknowledge the MIT Office of Research Computing and Data for providing high performance computing resources that have contributed to the research results reported within this paper.

The authors also acknowledge the use of Large Language Models (LLMs) in this work. In particular, ChatGPT (OpenAI; GPT-5.5 and GPT-5.6 Sol) was used throughout the preparation of this manuscript to review the text and suggest revisions to its writing and organization. All AI-assisted suggestions were reviewed and edited by the authors. ChatGPT was not used to generate experimental results or determine the scientific conclusions of the paper.


\bibliographystyle{IEEEtran}
\bibliography{references}

\appendices
\section{Model Problems}\label{appdx:ModelProblems}

Many of the constraints considered in this work are formulated in terms of conserved integral quantities. We therefore define the mass and energy of a solution $u$ at time $t$ by
\begin{align}
    m(t) &:= \int_{\Omega}u(x,t)\,dx \\
    E(t) &:= \frac12\int_{\Omega}u^2(x,t)\,dx
\end{align}

\subsection{Heat Equation Problems }
We first consider the one-dimensional heat equation
\begin{align}
    u_t = \alpha u_{xx}, \quad \Omega=[0,2\pi], \quad t \in [0,1]
\end{align}
with periodic boundary conditions $u(0,t) = u(2\pi, t)$, and initial values given by
\begin{align*}
    u(x,0) = u_{IC}^{\text{H}}(x):= \sin(x+\phi) 
\end{align*}
The spatiotemporal gridsizes are  $N_x = N_t = 100$.

For this specific problem, there is an analytical solution given by
\begin{align}
    u_{\text{analytic}}(x,t)  = \exp (-\alpha t)\sin(x+\phi)
\end{align}

The FFM model was trained using a dataset constructed by computing the analytical solution for different diffusion coefficients and phases. 
Namely, the parameters were sampled from the following distributions. 
\begin{align*}
\alpha\sim \mathcal{U}(1,5), \quad \phi \sim \mathcal{U}(0,\pi)
\end{align*}
The training dataset contained $10,000$ such solutions. 

During inference, we evaluate on two different constraint regimes.

In the first regime, we fix $\phi = \pi/4$ to keep the initial condition constant, 
and we enforce the initial value and mass conservation constraints.

\begin{align}
    \mathcal{H}^{\text{H,1}}(u) =
    \begin{bmatrix}
        u(x,0) - u_{IC}^{\text{H}}(x) \\
        m(t) - m(0)\\
    \end{bmatrix}
    \label{eq:heat1Constraint}
\end{align}

In the second set of constraints, we additionally impose an energy-evolution constraint together with a local PDE constraint. The resulting constraint operator is given by

\begin{align}
    \mathcal{H}^{\text{H,2}}(u) =
    \begin{bmatrix}
        u(x,0) - u_{IC}^{\text{H}}(x) \\
        m(t) - m(0)\\
        u_t -\alpha u_{xx}\\
        \frac{d}{dt} \frac12\int_{\Omega}(u(x,t))^2\;dx  +\alpha\int_{\Omega}(u_x)^2\;dx
    \end{bmatrix}
\end{align}

A derivation and motivation for the energy-evolution constraint are provided in Appendix \ref{appdx:constraintMotivation}.

\subsection{Burgers'}
Secondly, consider the one-dimensional inviscid Burgers' equation
\begin{align}
    u_t + \frac12(u^2)_x= 0, \quad \Omega = [0,1], \quad t\in[0,1]
\end{align}

The training dataset was constructed by generating solutions with the Godunov finite volume method \cite{leveque1992FVM}, on a $N_x = N_t = 101$ grid. For the training set, the following three conditions were imposed
\begin{enumerate}
    \item Initial value: \[
    u(x,0) = u_{\text{IC}}^{\text{B}}(x;p_{\text{loc}} ) :=\frac{1}{1+\exp(\frac{x-p_{\text{loc}}}{\epsilon})} 
    \]
    with $\epsilon = 0.02$ fixed, and the location sampled randomly from $p_{\text{loc}}\sim \mathcal{U}(0.2, 0.8)$. 
    \item Dirichlet condition on the left boundary
    \[
    u(0,t) = u_{\text{bc}} \quad \text{with } u_{\text{bc}}\sim \mathcal{U}(0,1)
    \]
    \item Neumann condition on the right 
    \[
    \frac{d}{dx}u(1,t) = 0
    \]
\end{enumerate}

$80$ different values of $p_{\text{loc}}$ and $u_{\text{bc}}$ were sampled each, generating a total of $6400$ solutions for the training dataset. 

The first regime of constraints considered for the Burgers' equation imposes a Dirichlet condition on the left boundary, and a Neumann condition on the right boundary. 
The Dirichlet value is not fixed, but varies across samples - as in the training data. 
There is also a mass conservation constraint. 
Together the constraints are given by 
\begin{align}
    \mathcal{H}^{\text{B,BC}}(u) =
    \begin{bmatrix}
        u(0, t) - u_L\\
        m(t) - m_0\\
        \partial_n u(1,t)
    \end{bmatrix}
    \label{eq:burgersBCconstraint}
\end{align}

In the second regime, we impose an initial value constraint, with $p_\text{loc}$ sampled as in the training set. 
In addition to the typical mass constraint we also impose a sequence of local conservation updates based on Godunov's flux method. 
The constraint function then becomes 

\begin{align}
    \mathcal{H}^{\text{B,IC}}(u) =
    \begin{bmatrix}
        u(x,0) - u_{\text{IC}}^{\text{B}}(x;p_{\text{loc}} )) \\
        m(t) - m(0)\\
        R^{(k)}_{\text{Flux}}(u) \; (k=1,..,5)
    \end{bmatrix}
    \label{eq:burgersICconstraint}
\end{align}
A derivation and description of the flux constraint based on the Godunov method are provided in Appendix \ref{appdx:constraintMotivation}.

\subsection{Reaction-Diffusion Equation}
Next, we consider a nonlinear reaction-diffusion equation given by 
\begin{align}
    u_t = \rho u (1-u) - \nu u_{xx} \\ \Omega = [0,1],\quad t\in [0,1]
\end{align}
with parameters $(\rho, \nu) = (0.01, 0.005)$. 
The equation is discretized on a grid with spatial and temporal resolutions $N_x = 128$ and $N_t = 100$, respectively.

The initial conditions are sampled from randomized combinations of sinusoidal and localized bump functions. The training dataset is generated by pairing $80$ distinct initial conditions with $80$ boundary conditions, resulting in a total of $6400$ PDE solutions. The solutions used for training are computed using a semi-implicit finite difference scheme.

\begin{align}
\mathcal{H}^{\text{RD}}(u)=
\begin{bmatrix}
u(x,0)-u_{\text{IC}}(x)
\\[0.3em]
\begin{aligned}
m(t)-\Bigl(
    m_0
    &+
    \int_0^{t}\rho u(1-u)\,d\tau \\
    &+
    \int_0^t
    (g_L(\tau)-g(\tau))\,d\tau
\Bigr)
\end{aligned}
\end{bmatrix}
\end{align}

\subsection{Navier-Stokes Equation}
Lastly we consider the Navier-Stokes equation in two dimensions, given in its vorticity form with periodic boundary conditions as 
\begin{align}
    w_t +u \cdot \nabla w = \nu \nabla^2w + f(x) \\ \Omega = [0,1]^2,\quad t\in [0,49]
\end{align}
where $w:=\nabla\times u$ is the vorticity, and we set the viscosity $\nu = 10^{-3}$. We let the forcing function $f(x) = 0.1\sqrt{2}\sin(2\pi(x+y)+\phi)$. The spatiotemporal resolutions are as follows
$N_x =N_y = 16$, $N_t = 50$.
The training dataset was constructed by solving the governing problem with a Crank–Nicolson spectral solver. Specifically, $10,000$ simulations were generated from  $100$ randomly sampled initial vorticities $w_0$, drawn from a Gaussian random field, and $100$ forcing phases $\phi \sim \mathcal{U}(0, \pi/2)$. With periodic boundary conditions the vorticity-mass is conserved. We further impose an initial value constraint, and an energy conservation constraint on the vorticity. In total, these are the constraints:

\begin{align}
  \mathcal{H}^{\text{NS}}(w) =
  \begin{bmatrix}
  w(x,y,0) - w_{\text{IC}}(x,y) \\[0.4em]
  \displaystyle\int_{\Omega} w(x,y,t)\,dx\,dy -
  \int_{\Omega} w_{\text{IC}}(x,y)\,dx\,dy \\[0.4em]
  \displaystyle\int_{\Omega} w(x,y,t)^2\,dx\,dy -
  \int_{\Omega} w_{\text{IC}}(x,y)^2\,dx\,dy
  \end{bmatrix}
\end{align}
\section{Newton-KKT system}\label{appdx:newtonKKT}
Consider the equality constrained optimisation problem 

\begin{align}
    \min_{x\in \mathbb{R}^n} f(x)\quad \text{subject to } g(x) =0
    \label{eq:constrainedOpt}
\end{align}

for some objective function $f: \mathbb{R}^n\to \mathbb{R}$, and constraint function $g:\mathbb{R}^n\to \mathbb{R}^m$

Our goal is to show that approximating this solution can be reduced to repeatedly solving linear systems
of the form 
$Au = b$, where $A$ has a characteristic sparse structure that is highly dependent on the Jacobian 
of $g$.

The key idea to solving \eqref{eq:constrainedOpt} is that at the optimum, 
we cannot improve the objective while remaining in the feasible set. To illustrate this, suppose $x^*$ solves \eqref{eq:constrainedOpt}. Consider a small perturbation 
$\Delta x$. To remain within the feasible set we must require 
\[
g(x^* + \Delta x) = 0
\]
Linearization gives \[g(x^*) + J(x^*)\Delta x = 0\]
where we have introduced the Jacobian of $g$, $$J(u):= \nabla_x g(u) \in \mathbb{R}^{m\times n}$$
As $x^*$ by assumption is in the feasible set, we have that $J(x^*)\Delta x = 0$. 
In other words, all feasible perturbations must lie in the nullspace of the Jacobian. 

Moreover, as $x^*$ is an optimum, one cannot decrease the objective function in any 
feasible direction. This implies that 
\[
\nabla f(x^*)^T \Delta x = 0, \quad \forall \Delta x \text{ satisfying } J(x^*)\Delta x = 0
\]

A basic linear algebra result is that a vector $v$ which is orthogonal to the nullspace of a matrix 
$Q$, is necessarily in the rowspace of $Q$. 
This implies that we can write 
\[
\nabla f(x^*) = - J(x^*)^T\lambda
\]
for some $\lambda \in \mathbb{R}^m$. 

These two conditions together form the Karush-Kuhn-Tucker (KKT) conditions that 
we require the solution to satisfy

\begin{align*}
    \nabla f(x) + (J(x))^T \lambda  &= 0\\
        g(x) &= 0 
\end{align*}
By introducing the Lagrangian $\mathcal{L}(x,\lambda) = f(x) + (J(x))^T\lambda$ 
we can also reformulate the problem of finding $x^*$ so that 
$F(x, \lambda) = 0$, where we define  

\begin{align}
   F:\mathbb{R}^{n+m}\to \mathbb{R}^{n+m}, \quad  F(x, \lambda) := \begin{bmatrix}
        \nabla_x \mathcal{L}(x,\lambda)  \\
        g(x) 
    \end{bmatrix}
\end{align}

One can recall that a single iteration of the multivariate Newton-Raphson 
method for solving the problem $$G(u):\mathbb{R}^p \to \mathbb{R}^q$$
takes the form 
\begin{align*}
    u^{(k+1)} = u^{(k)} - (\nabla G(u^{(k)}))^{-1}G(u^{(k)})
\end{align*}
which can be simplified to solving the linear equation 
\begin{align}
    \nabla G(u^{(k)}) \Delta_k  =  -G(u^{(k)}), \quad \Delta_k := u^{(k+1)} - u^{(k)}
\end{align}

Hence, to solve $F(x,\lambda) =0$ using the Newton-Raphson method, 
we need to compute $\nabla_{(x,\lambda)}F(x,\lambda)$. 

A straightforward computation yields the linear system 
\begin{align}
    \begin{bmatrix}
        W & J^T\\
        J & 0 
    \end{bmatrix}
     \begin{bmatrix}
        \Delta x_n \\
        \Delta \lambda_n
     \end{bmatrix}
      &= 
      -\begin{bmatrix}
        \nabla f(x_n) + (J(x_n))^T \lambda_n \\
        g(x_n)
      \end{bmatrix}
       \\
      W &:= \nabla_{xx}^2 \mathcal{L}(x_n,\lambda_n)
      \label{eq:newtonKKTsystem}
\end{align}

The matrix on the left-hand side of \eqref{eq:newtonKKTsystem} is called the \textit{Newton-KKT system}. Its sparse block structure plays a central role in large-scale constrained optimisation algorithms.
\section{Jacobian Structure Example}\label{appdx:toyConstraints}
This appendix presents an example illustrating the sparsity structure in the Jacobian arising from a common set of PDE constraints. 

Consider a PDE subject to the following constraints. 
\begin{align*}
    \mathcal{H}(u) = \begin{bmatrix}
        u(x,0) - f(x)\quad \forall x\in \Omega\\
        \int_{\Omega}u(x,t)\;dx - \int_{\Omega}u(x,0)\;dx\quad  \forall t\in [0,T]\\
    \end{bmatrix}
    = \mathbf{0}
\end{align*}
Assume the solution is discretized on a spatiotemporal grid with $(N_x, N_t) = (3,3)$ grid points: $x_1, x_2, x_3$, and $t_1,t_2,t_3$. Let $\mathbf{u}_i^{(k)}$ denote the approximation of $u(x_i, t_k)$. 

The initial value constraint then becomes 
\begin{align*}
    \mathbf{u}_i^{(1)} - f_i = 0, \qquad f_i:=f(x_i), \quad i=1,2,3
\end{align*}
Using a simple left Riemann sum approximation to the integral, the mass conservation constraint takes the form 
\begin{align*}
    \sum_{i=1}^2 \mathbf{u}_i^{(k)} - \sum_{i=1}^2 \mathbf{u}_i^{(1)} =0, \quad k = 2, 3
\end{align*}
where the constraint at the first timestep is omitted. 

In total, this yields $5$ constraints. For convenience, we represent the approximation $\mathbf{u}$ as the 
flattened vector \[\mathbf{u} = (\mathbf{u}_1^{(1)}, \mathbf{u}_2^{(1)}, \mathbf{u}_3^{(1)}, ..., \mathbf{u}_1^{(3)}, \mathbf{u}_2^{(3)}, \mathbf{u}_3^{(3)})^T\in \mathbb{R}^9\].

The Jacobian of the constraint operator, $J_{\mathcal{H}}\in \mathbb{R}^{5\times9}$ has entries 
$(J_{\mathcal{H}})_{i,j} = \frac{\partial \mathcal{H}_i}{\partial \mathbf{u}_j}$, where the constraints are ordered such that the initial value constraints appear first.

A straightforward computation gives 

\begin{align}
J_{\mathcal{H}} = 
    \begin{bmatrix}
        \textbf{1} & 0 & 0 & 0 & 0 & 0 & 0 & 0 & 0 \\
        0 & \textbf{1} & 0 & 0 & 0 & 0 & 0 & 0 & 0 \\
        0 & 0 & \textbf{1} & 0 & 0 & 0 & 0 & 0 & 0 \\
       \textbf{-1}&\textbf{-1}& 0 & \textbf{1}& \textbf{1} & 0 & 0 & 0 & 0 \\
      \textbf{-1} & \textbf{-1}&0 & 0 & 0 & 0 & \textbf{ 1}&  \textbf{1}& 0 \\
    \end{bmatrix}
\end{align}

The $3 \times 3$ identity block in the upper-left corner arises from the initial value constraints.
More generally, initial value constraints typically contribute block-diagonal structure to the Jacobian.

Similarly, regardless of the number of grid points, a mass conservation constraint discretized using a left Riemann sum produces rows containing a sequence of $N_x - 1$ nonzero entries followed by a zero, reflecting the structure of the underlying quadrature rule.
\section{Constraints}\label{appdx:constraintMotivation}

\subsection{Energy Evolution Constraint}
In generating solutions for the heat equation we considered an energy evolution constraint. Here, "energy" is meant as \[
E(t) :=\frac12\int_{\Omega}(u(x,t))^2\;dx
\]
not to be confused with the physical energy of the system. The energy referred to here is more of a measure of variance of $u$. We might want to enforce that the temperature profile, $u$ is smoothed out over time. In other words, that the temporal derivative $
\frac{d}{dt}E(t)\leq 0$. If one assumes zero flux on the boundary, then this is a purely mathematical artifact of the heat equation. 
\begin{align*}
    \frac{d}{dt} \frac12\int_{\Omega}(u(x,t))^2\;dx & = \int_{\Omega} u u_t \; dx \\
    = \int_{\Omega} \alpha uu_{xx}\; dx &=-\alpha\int_{\Omega}(u_x)^2\;dx
\end{align*}
where the second-to-last equality comes from substituting in the heat equation, and the last equality comes from integration by parts - assuming zero flux on boundary. Importantly, the last expression must be negative, hence, we can enforce 
\[
\frac{d}{dt} \frac12\int_{\Omega}(u(x,t))^2\;dx  = -\alpha\int_{\Omega}(u_x)^2\;dx
\]
This constraint enforces that the energy of the solution, $E(t)$, which is to 
be interpreted as the variability of the solution in space, decreases over time - much as one would expect from a diffusive process such as the heat equation.

\subsection{Godunov's Flux Conservation}

For the Burgers' equation with fixed initial condition, we additionally imposed a sequence of local conservation constraints based on Godunov's finite-volume method. The purpose of this constraint was to encourage the neural solution to satisfy not only global mass conservation, but also the local conservative transport structure of the PDE.

Specifically, the constraint operator included the residual terms
\[
R^{(k)}_{\text{Flux}}(u), \qquad k=1,\dots,5,
\]
where each residual corresponds to one conservative update step of a finite-volume discretization.

Let the spatial domain be partitioned into cells indexed by $i$, with cell averages
\[
u_i^k \approx \frac{1}{\Delta x}\int_{x_i}^{x_{i+1}} u(x,t_k)\,dx.
\]
For Burgers' equation
\[
u_t + \frac12(u^2)_x = 0,
\]
Godunov's method updates the solution according to
\[
u_i^{k+1}
=
\nu_i^k
-
\frac{\Delta t}{\Delta x}
\left(
F_{i+\frac12}^k - F_{i-\frac12}^k
\right),
\]
where $F_{i+\frac12}^k$ denotes the numerical flux through the interface between neighboring cells.

For Burgers' equation, the Godunov flux is given by
\[
F(a,b)
=
\begin{cases}
\min_{u\in[a,b]} \frac{u^2}{2},
& a \le b,
\\[0.3em]
\max_{u\in[b,a]} \frac{u^2}{2},
& a > b,
\end{cases}
\]
with $a=u_i^k$ and $b=u_{i+1}^k$.

Using the neural network prediction evaluated on the spatial grid, we defined the flux residual as the mismatch between consecutive predicted states and a single Godunov update:
\[
R_{\text{Flux}}^{(k)}(u)
=
u^{k+1}
-
\left[
u^k
-
\frac{\Delta t}{\Delta x}
\left(
F_{i+\frac12}^k - F_{i-\frac12}^k
\right)
\right].
\]

Thus, the constraint enforces that the predicted solution approximately evolves according to a locally conservative finite-volume scheme over several successive time steps.

\subsection{Discretization}

To evaluate the conservation constraints in practice, the continuous space-time domain was discretized on a uniform grid
\[
x_i = i\Delta x,
\qquad
t_k = k\Delta t,
\]
where $\Delta x$ and $\Delta t$ denote the spatial and temporal step sizes, respectively. Using the representation that $\mathbf{u}_i^k$ is the approximation to $u(x_i, t_k)$, the continuous conservation operators could be approximated through discrete sums and local update relations.

Global conservation quantities were approximated through Riemann sums over the spatial grid. For example, the total mass at time $t_k$ was approximated as
\[
m(t_k)
\approx
\sum_i u_i^k \Delta x.
\]

\subsubsection{Neumann Boundary Conditions}

For problems with homogeneous Neumann boundary conditions, the continuous condition
\[
\frac{\partial u}{\partial n} = 0
\]
was enforced discretely by requiring the normal derivative at the boundary to vanish. In one spatial dimension, this corresponds to
\[
\frac{u_2^k - u_1^k}{\Delta x} = 0,
\qquad
\frac{u_{N_x}^k - u_{N_x-1}^k}{\Delta x} = 0,
\]
which implies that the boundary values mirror their neighboring interior values,
\[
u_1^k = u_2^k,
\qquad
u_{N_x}^k = u_{N_x-1}^k.
\]

This prevents artificial fluxes through the boundary and ensures consistency with the conservation constraints.

\subsubsection{Discretization of the Conservation Constraint}

Consider the continuous conservation constraint
\[
\frac{d}{dt}\int_\Omega \rho(u)\,dx - C = 0,
\]
where $\rho(u)$ denotes some quantity whose integral in space we'd like to conserve,  and $C$ is some constant.

Using the spatial discretization, the integral term was approximated through a Riemann sum,
\[
\int_\Omega \rho(u)\,dx
\approx
\sum_i \rho(u_i^k)\Delta x.
\]

The temporal derivative was then approximated using a finite difference,
\[
\frac{d}{dt}\int_\Omega \rho(u)\,dx
\approx
\frac{
\sum_i \rho(u_i^{k+1})\Delta x
-
\sum_i \rho(u_i^{k})\Delta x
}{\Delta t}.
\]

Hence, the fully discretized conservation constraint becomes
\[
\frac{
\sum_i \rho(u_i^{k+1})\Delta x
-
\sum_i \rho(u_i^{k})\Delta x
}{\Delta t}
- C.
\]

\section{Additional Results}\label{appdx:additionalResults}

\subsection{Constraint Violation}
\label{app:constraint_violation}

To verify that the optimization methods formulated as constrained nonlinear
programs actually enforce the imposed physical constraints, we evaluate the
residual of each constraint separately on the generated samples. Let
$\hat{u}^{(n)}$ denote the final generated solution for sample
$n\in\{1,\ldots,N_s\}$, and let
$\mathcal{R}_{k}(\hat{u}^{(n)})$ denote the residual associated with constraint
type $k$. Examples include initial-value conditions, boundary conditions,
mass conservation, energy evolution, and local flux constraints.

For each constraint type, we report the mean Euclidean norm of the residual
over the generated batch:
\begin{equation}
    \text{Constraint Error}_k
    =
    \frac{1}{N_s}
    \sum_{n=1}^{N_s}
    \left\|
        \mathcal{R}_{k}\!\left(\hat{u}^{(n)}\right)
    \right\|_2 .
    \label{eq:constraint_error}
\end{equation}
Here, $N_s$ is the number of generated samples and $E_k$ is the reported
constraint error for constraint type $k$. Each constraint is evaluated using
the same discretization employed in the corresponding projection problem.

The purpose of this metric is to confirm that the constrained optimization
methods solve the stated projection problem to numerical precision. It is not
intended to rank different constraint types, since their residual vectors may
have different dimensions and physical scales. Comparisons should therefore
be made between optimization methods for the same problem and constraint.

Tables~\ref{tab:CE_heat},~\ref{tab:CE_heat2},~\ref{tab:CE_burgersBC},~\ref{tab:CE_rd},~\ref{tab:CE_ns}, and ~\ref{tab:CE_burgers_IC_Mass_Flux_5}
report the constraint errors for all benchmark
problems. A separate column is provided for each imposed constraint, allowing
the feasibility of initial conditions, boundary conditions, conservation laws,
and local PDE constraints to be inspected individually. A dash indicates that
the corresponding method did not complete within the allotted runtime budget
and therefore produced no solution for evaluation.

Across the benchmark problems, the methods that explicitly solve the
equality-constrained nonlinear program---\texttt{ExaModels + MadNLP} on GPU
and CPU, \texttt{JuMP + MadNLP}, and \texttt{JuMP + Ipopt}---produce residuals
at or near numerical precision. These results confirm that the runtime
differences reported in the main text are not obtained by relaxing or
neglecting the imposed constraints. In contrast, \texttt{IPNewton} and
\texttt{L-BFGS} do not consistently attain the same level of feasibility.
This is expected for \texttt{L-BFGS}, which treats constraint residuals
through a penalty objective rather than imposing them as hard equality
constraints.

\begin{table}[htbp]
\caption{Constraint Errors: Heat - 1}
\label{tab:CE_heat}
\begin{center}
\begin{tabular}{|c|c|c|}
\hline
\textbf{Optimization}&\multicolumn{2}{|c|}{\textbf{Constraint Error}} \\
\cline{2-3} 
\textbf{Setup} & \textbf{\textit{IC}}& \textbf{\textit{Mass}} \\
\hline
Exa, MadNLP, GPU & 7.66E-07& 6.59E-06 \\
Exa, MadNLP, CPU & 1.11E-06&  9.25E-06 \\
JuMP, MadNLP, CPU & 7.44E-07 &  1.36E-05 \\
JuMP, Ipopt, CPU & 7.44E-07 &  1.35E-05 \\
IPNewton & 1.38E-06 &   2.606\\
L-BFGS & 1.41E-06  & 2.246 \\
\hline
\end{tabular}
\label{tab1}
\end{center}
\end{table}

\begin{table}[htbp]
\caption{Constraint Errors: Heat - 2}
\label{tab:CE_heat2}
\begin{center}
\begin{tabular}{|c|c|c|c|c|}
\hline
\textbf{Optimization}&\multicolumn{4}{|c|}{\textbf{Constraint Error}} \\
\cline{2-5} 
\textbf{Setup} & \textbf{\textit{IC}}& \textbf{\textit{Mass}}& \textbf{\textit{PDE}}& \textbf{\textit{Energy}} \\
\hline
Exa, MadNLP, GPU &2.56E-06 & 2.06E-06 &4.33E-05 & 3.73E-05\\
Exa, MadNLP, CPU &7.44E-07 & 3.59E-07 &4.06E-05 &0\\
JuMP, MadNLP, CPU & - & - & - & -\\
JuMP, Ipopt, CPU & - & - & - & - \\
IPNewton & - & - & - & - \\
L-BFGS & 1.41E-06 & 4.75E-04  & 1.37E-02 & 0\\
\hline
\end{tabular}
\label{tab1}
\end{center}
\end{table}

\begin{table}[htbp]
\caption{Constraint Errors: Burgers' (BC)}
\label{tab:CE_burgersBC}
\begin{center}
\begin{tabular}{|c|c|c|c|}
\hline
\textbf{Optimization}&\multicolumn{3}{|c|}{\textbf{Constraint Error}} \\
\cline{2-4} 
\textbf{Setup} & \textbf{\textit{Dirichlet}}& \textbf{\textit{Neumann}}&  \textbf{\textit{Mass}} \\
\hline
Exa, MadNLP, GPU & 0 & 0  & 7.01E-07 \\
Exa, MadNLP, CPU & 0 & 0  & 6.65E-07 \\
JuMP, MadNLP, CPU& 0 & 0  & 6.63E-07 \\
JuMP, Ipopt, CPU & 0 & 0  & 6.63E-07 \\
IPNewton    & - & -  & -        \\
L-BFGS & 5.54E-05& 7.75E-06  & 0.03467\\
\hline
\end{tabular}
\label{tab1}
\end{center}
\end{table}

\begin{table}[htbp]
\caption{Constraint Errors: Reaction-Diffusion Equation}
\label{tab:CE_rd}
\begin{center}
\begin{tabular}{|c|c|c|}
\hline
\textbf{Optimization}&\multicolumn{2}{|c|}{\textbf{Constraint Error}} \\
\cline{2-3} 
\textbf{Setup} & \textbf{\textit{IC}}& \textbf{\textit{Mass}} \\
\hline
Exa, MadNLP, GPU &1.41E-06 & 6.47E-05 \\
Exa, MadNLP, CPU &6.27E-07 &  9.17E-07 \\
JuMP, MadNLP, CPU & 1.70E-18 &  9.01E-07 \\
JuMP, Ipopt, CPU & 2.82E-32 & 9.01E-07  \\
IPNewton & 8.64E-07 &  4.61E-03 \\
L-BFGS & 8.37E-07  &  4.47E-03\\
\hline
\end{tabular}
\label{tab1}
\end{center}
\end{table}

\begin{table}[htbp]
\caption{Constraint Errors: Navier-Stokes' Equation}
\label{tab:CE_ns}
\begin{center}
\begin{tabular}{|c|c|c|c|}
\hline
\textbf{Optimization}&\multicolumn{3}{|c|}{\textbf{Constraint Error}} \\
\cline{2-4} 
\textbf{Setup} & \textbf{\textit{IC}}& \textbf{\textit{Mass}}& \textbf{\textit{Enstrophy}} \\
\hline
Exa, MadNLP, GPU & 1.70E-10 & 2.01E-05 & 6.83E-07 \\
Exa, MadNLP, CPU & 1.29E-06 &9.94E-06& 5.96E-07 \\
JuMP, MadNLP, CPU & 0 & 8.89E-05 & 4.18E-07\\
JuMP, Ipopt, CPU & 0&  8.89E-05& 4.18E-07\\
IPNewton & -  &  - & - \\
L-BFGS & 2.647 & 3.16E-04 &  2.49E-01\\
\hline
\end{tabular}
\label{tab1}
\end{center}
\end{table}

\begin{table}[htbp]
\caption{Constraint Errors: Burgers' (IC)}
\label{tab:CE_burgers_IC_only}
\begin{center}
\begin{tabular}{|c|c|}
\hline
\textbf{Optimization}&\multicolumn{1}{|c|}{\textbf{Constraint Error}} \\
\cline{2-2} 
\textbf{Setup} & \textbf{\textit{IC}} \\
\hline
Exa, MadNLP, GPU & 8.09E-07 \\
Exa, MadNLP, CPU &  8.67E-07 \\
JuMP, MadNLP, CPU &  3.69E-07 \\
JuMP, Ipopt, CPU &    3.69E-07 \\
IPNewton &    1.16E-06 \\
L-BFGS &    1.17E-06 \\
\hline
\end{tabular}
\label{tab1}
\end{center}
\end{table}

\begin{table}[htbp]
\caption{Constraint Errors: Burgers' (IC, Mass)}
\label{tab:CE_burgers_IC_Mass}
\begin{center}
\begin{tabular}{|c|c|c|}
\hline
\textbf{Optimization}&\multicolumn{2}{|c|}{\textbf{Constraint Error}} \\
\cline{2-3} 
\textbf{Setup} & \textbf{\textit{IC}}& \textbf{\textit{Mass}} \\
\hline
Exa, MadNLP, GPU & 3.69E-07 & 4.17E-07 \\
Exa, MadNLP, CPU & 8.67E-07&  4.85E-07 \\
JuMP, MadNLP, CPU & 3.69E-07 &  6.77E-07 \\
JuMP, Ipopt, CPU & - & -  \\
IPNewton &  - &  - \\
L-BFGS &  - & -  \\
\hline
\end{tabular}
\label{tab1}
\end{center}
\end{table}

\begin{table}[htbp]
\caption{Constraint Errors: Burgers' (IC, Mass, Flux ($k=1$))}
\label{tab:CE_burgers_IC_Mass_Flux_1}
\begin{center}
\begin{tabular}{|c|c|c|c|}
\hline
\textbf{Optimization}&\multicolumn{3}{|c|}{\textbf{Constraint Error}} \\
\cline{2-4} 
\textbf{Setup} & \textbf{\textit{IC}}& \textbf{\textit{Mass}}& \textbf{\textit{Flux $(k=1)$}} \\
\hline
Exa, MadNLP, GPU & 1.02E-06& 9.20E-06 & 9.61E-07 \\
Exa, MadNLP, CPU & 8.67E-07 & 4.87E-07 & 9.26E-07 \\
JuMP, MadNLP, CPU & - & - & - \\
JuMP, Ipopt, CPU & - & -  & - \\
IPNewton & -  & - & - \\
L-BFGS & - &  - & - \\
\hline
\end{tabular}
\label{tab1}
\end{center}
\end{table}

\begin{table}[htbp]
\caption{Constraint Errors: Burgers' (IC, Mass, Flux ($k=5$))}
\label{tab:CE_burgers_IC_Mass_Flux_5}
\begin{center}
\begin{tabular}{|c|c|c|c|}
\hline
\textbf{Optimization}&\multicolumn{3}{|c|}{\textbf{Constraint Error}} \\
\cline{2-4} 
\textbf{Setup} & \textbf{\textit{IC}}& \textbf{\textit{Mass}}& \textbf{\textit{Flux $(k=5)$}} \\
\hline
Exa, MadNLP, GPU & 6.63E-06& 2.74E-05 & 1.23E-05\\
Exa, MadNLP, CPU & 9.82E-07 & 6.10E-07 & 2.47E-06\\
JuMP, MadNLP, CPU & - & - & - \\
JuMP, Ipopt, CPU & - & -  & - \\
IPNewton & -  & - & - \\
L-BFGS & - &  - & - \\
\hline
\end{tabular}
\label{tab1}
\end{center}
\end{table}

\begin{table}[htbp]
\caption{Constraint Errors: Burgers' (IC, Mass, Flux ($k=10$))}
\label{tab:CE_burgers_IC_Mass_Flux_10}
\begin{center}
\begin{tabular}{|c|c|c|c|}
\hline
\textbf{Optimization}&\multicolumn{3}{|c|}{\textbf{Constraint Error}} \\
\cline{2-4} 
\textbf{Setup} & \textbf{\textit{IC}}& \textbf{\textit{Mass}}& \textbf{\textit{Flux $(k=10)$}} \\
\hline
Exa, MadNLP, GPU & 8.20E-06 & 3.34E-05 & 2.54E-05\\
Exa, MadNLP, CPU & - & - & - \\
JuMP, MadNLP, CPU & - & - & - \\
JuMP, Ipopt, CPU & - & -  & - \\
IPNewton & -  & - & - \\
L-BFGS & - &  - & - \\
\hline
\end{tabular}
\label{tab1}
\end{center}
\end{table}

\subsection{Scaling Study - Number of Samples}

\begin{figure}[h]
  \centering
  \includegraphics[width=\linewidth]{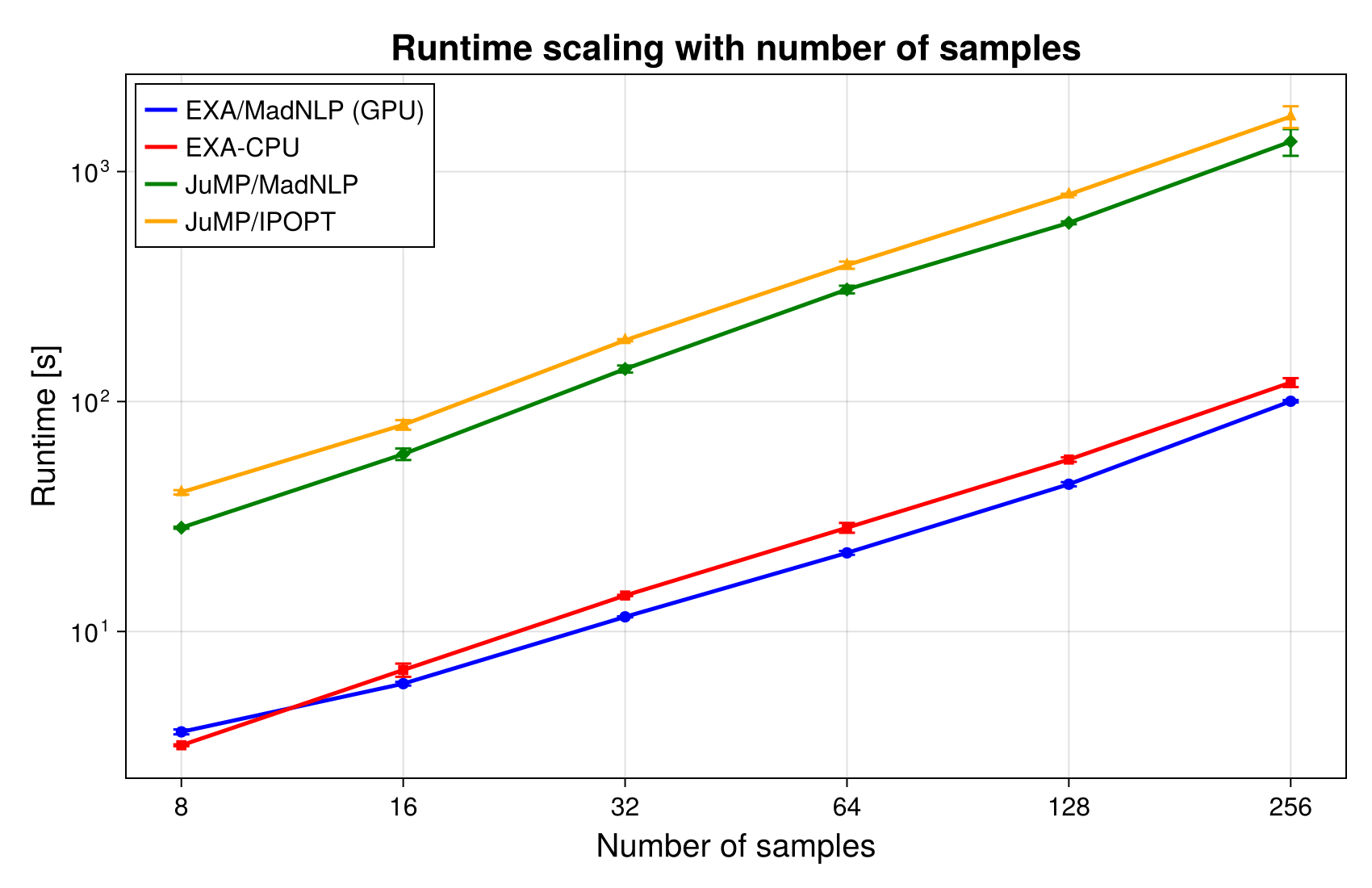}
 \caption{Runtime scaling with the number of generated samples for the first heat-equation constraint regime. Time scales linearly with number of samples, and ordering is preserved.}
  \label{fig:nsample_scale}
\end{figure}

Figure \ref{fig:nsample_scale} shows the runtime of each optimization backend as the number of generated samples is increased. The runtime grows approximately linearly with the number of samples for all methods. This is consistent with the block-diagonal structure of the batched projection problem: increasing the batch size adds independent constraint blocks, but does not introduce additional cross-sample coupling. The relative ordering of the methods is also largely preserved across batch sizes.



\subsection{Scaling Study - Number of Constraints}

Table~\ref{tab:scaleStudyConstrainsTime} shows the runtime as additional constraints are imposed on the Burgers equation. Even for the smaller nonlinear constraint sets, MadNLP-based methods are substantially faster than the alternative baselines. As local flux constraints are added, the projection problem becomes significantly more difficult. In this regime, GPU acceleration becomes increasingly important. The ExaModels + MadNLP GPU backend is the only method that remains tractable for the largest tested constraint set.

 \textit{IC-only} fixes the initial condition, while \textit{IC, Mass} additionally enforces conservation of the spatial integral over time. The Flux($k$) constraints add $k$ local Godunov update residuals -- see Appendix \ref{appdx:constraintMotivation}.

Tables~\ref{tab:CE_burgers_IC_only},~\ref{tab:CE_burgers_IC_Mass}, ~\ref{tab:CE_burgers_IC_Mass_Flux_1}, ~\ref{tab:CE_burgers_IC_Mass_Flux_5}, and ~\ref{tab:CE_burgers_IC_Mass_Flux_10} show the constraint violations.

\begin{table*}[t]
  \caption{Runtime scaling with the number of imposed constraints for the Burgers equation.}
  \label{tab:scaleStudyConstrainsTime}
  \begin{center}
    \begin{small}
      \begin{sc}
        \begin{tabular}{lccccc}
          \toprule
          Optimization Setup & IC-only & IC, Mass & IC, Mass, Flux(1) & IC, Mass, Flux(5) & IC, Mass, Flux(10) \\
          \midrule
          Exa, MadNLP, GPU  & 6.90$\pm$0.49 & 19.32$\pm$0.49 & 24.23$\pm$0.09 & 39.03$\pm$0.36 & 62.95$\pm$2.07 \\
          Exa, MadNLP, CPU  & 25.56$\pm$0.17 & 24.40$\pm$0.51 & 1632.12$\pm$73.94 & 2038.96$\pm$215.10 & DNF \\
          JuMP, MadNLP, CPU & 155.33$\pm$1.61 & 346.27$\pm$3.75 & DNF & DNF & DNF \\
          JuMP, Ipopt, CPU  & 156.28$\pm$1.87 & DNF & DNF & DNF & DNF \\
          IPNewton          & 50.38$\pm$3.56 & DNF & DNF & DNF & DNF \\
          L-BFGS            & 3.92$\pm$0.01 & DNF & DNF & DNF & DNF \\
          \bottomrule
        \end{tabular}
      \end{sc}
    \end{small}
  \end{center}
  \vskip -0.1in
\end{table*}

\section{Implementation Details}\label{appdx:implementationDetails}

All sampling runs were initialized from white noise -- $iid$ Gaussian entries -- matched to the grid of the target PDE as a starting point for our models. 

For all the methods, the forward pass of the pretrained neural operator is executed on the GPU. This choice is intentional to isolate the cost of the projection step, rather than benchmark CPU versus GPU neural-network inference. A fully CPU-resident implementation would likely incur additional cost from evaluating the neural operator on the CPU, and should therefore be expected to have larger end-to-end runtimes than those reported here. 

The optimization models and associated problem data were constructed
using single-precision floating-point values (\texttt{Float32}).
To ensure a consistent comparison across constrained nonlinear optimization methods, the
solver convergence tolerance was fixed to $10^{-5}$ for both
\texttt{MadNLP} and \texttt{Ipopt}. Thus, the reported constraint residuals
should be interpreted relative to the prescribed solver tolerance and the
numerical precision of the \texttt{Float32} formulation.

\subsection{Hardware}
All experiments were conducted on a high-performance computing cluster with 1 × NVIDIA L40S GPU (46 GB VRAM), 4 CPU cores, and 64 GB RAM.

\end{document}